\documentclass[10pt]{article} 

\PassOptionsToPackage{normalem}{ulem}
\PassOptionsToPackage{noend}{algpseudocode}
\PassOptionsToPackage{suppress}{color-edits}

\usepackage{etoolbox}
\usepackage{comment}
\newtoggle{neurips}
\togglefalse{neurips}

\newcommand{\arxiv}[1]{\iftoggle{neurips}{}{#1}}

\usepackage[letterpaper, left=.95in, right=.95in, top=.95in,bottom=.95in]{geometry}
\usepackage{parskip}
\usepackage{amsmath,amsthm,amssymb}
\usepackage{mathtools,bbm,xspace}
\usepackage{braket}
\usepackage{thm-restate}
\usepackage{algorithm}
\usepackage{algpseudocode}

\usepackage[pagebackref]{hyperref}
\usepackage[nameinlink,capitalise]{cleveref}

\usepackage[dvipsnames]{xcolor}
\hypersetup{
colorlinks=true,
urlcolor=blue,
linkcolor=RoyalBlue,
citecolor=PineGreen,
}
\usepackage{ulem}
\usepackage[shortlabels]{enumitem}
\usepackage{multirow}

\usepackage{bm}
\usepackage{csquotes}
\allowdisplaybreaks

\usepackage{graphicx}

\newtheorem*{theorem*}{Theorem}

\newtheorem*{proposition*}{Proposition}

\newtheorem*{lemma*}{Lemma}

\newtheorem*{conjecture*}{Conjecture}

\newtheorem*{fact*}{Fact}

\newtheorem*{exercise*}{Exercise}

\newtheorem*{hypothesis*}{Hypothesis}

\theoremstyle{definition}

\newtheorem{exercise-easy}[theorem]{Exercise}
\newtheorem{exercise-med}[theorem]{Exercise}
\newtheorem{exercise-hard}[theorem]{Exercise$^\star$}

\newtheorem*{claim*}{Claim}

\newtheorem*{remark*}{Remark}

\newtheorem*{observation*}{Observation}

\newcommand{\mylabel}[2]{\def\@currentlabel{#2}\label{#1}}

\newcommand{\Sref}[1]{\hyperref[#1]{\S\ref*{#1}}}

\newcommand\whichfont{0}
\ifnum\whichfont=1

\else
\newcommand\isBB{0}
\ifnum\isBB=0

\else

\fi
\fi

\newcommand\MYcurrentlabel{xxx}

\newcommand{\MYstore}[2]{%
  \global\expandafter \def \csname MYMEMORY #1 \endcsname{#2}%
}

\newcommand{\MYload}[1]{%
  \csname MYMEMORY #1 \endcsname%
}

\newcommand{\MYnewlabel}[1]{%
  \renewcommand\MYcurrentlabel{#1}%
  \MYoldlabel{#1}%
}

\newcommand{\MYdummylabel}[1]{}

\newcommand{\torestate}[1]{%
  \let\MYoldlabel\label%
  \let\label\MYnewlabel%
  #1%
  \MYstore{\MYcurrentlabel}{#1}%
  \let\label\MYoldlabel%
}
\newcommand{\restateprop}[1]{%
  \let\MYoldlabel\label
  \let\label\MYdummylabel
  \begin{proposition*}[Restatement of \pref{#1}]
    \MYload{#1}
  \end{proposition*}
  \let\label\MYoldlabel
}
\newcommand{\restatelemma}[1]{%
  \let\MYoldlabel\label
  \let\label\MYdummylabel
  \begin{lemma*}[Restatement of \pref{#1}]
    \MYload{#1}
  \end{lemma*}
  \let\label\MYoldlabel
}
\newcommand{\restateclaim}[1]{%
  \let\MYoldlabel\label
  \let\label\MYdummylabel
  \begin{claim*}[Restatement of \pref{#1}]
    \MYload{#1}
  \end{claim*}
  \let\label\MYoldlabel
}

\newcommand{\restatetheorem}[1]{%
  \let\MYoldlabel\label
  \let\label\MYdummylabel
  \begin{theorem*}[Restatement of \pref{#1}]
    \MYload{#1}
  \end{theorem*}
  \let\label\MYoldlabel
}

\newcommand{\bbE}{\mathbb E}

\newcommand{\bx}{{\boldsymbol{x}}}
\newcommand{\by}{\boldsymbol{y}}

\renewcommand{\geq}{\geqslant}

\renewcommand{\emph}{\textit}

\usepackage{bm}

\usepackage{subcaption}
\usepackage{graphicx}
\graphicspath{{images/}}

\usepackage{tikz}
\usetikzlibrary{calc,positioning}
\usetikzlibrary{decorations.pathreplacing}
\usepackage{scalerel}

\newcommand{\Lsft}{\mathcal{L}_{\mathrm{SFT}}}
\newcommand{\Lrl}{\mathcal{L}_{\mathrm{RL}}}
\newcommand{\pibase}{\pi_{\mathrm{base}}}
\newcommand{\pisft}{\pi_{\mathrm{SFT}}}

\newcommand{\Dsft}{D_{\mathrm{SFT}}}
\newcommand{\Drl}{D_{\mathrm{RL}}}

\usepackage{cases}

\newcommand{\cX}{\mathcal{X}}
\newcommand{\cY}{\mathcal{Y}}

\newcommand{\Dval}{D_{\mathrm{val}}}

\usepackage{xcolor}



\usepackage[most]{tcolorbox}
\tcbuselibrary{skins,breakable}

\newtcolorbox[
  auto counter,
  number within=section,
  number freestyle={\noexpand\thesection.\noexpand\arabic{\tcbcounter}~\noexpand\mytitle},
  crefname={Meta-algorithm}{Meta-algorithms}
]{metaalgorithm}[2][]{
  breakable,
  enhanced,
  colback=white,
  colframe=black,
  coltitle=black,
  fonttitle=\bfseries,
  code={\def\mytitle{#2}},     
  title=Meta-algorithm \thetcbcounter,
  boxrule=0.8pt,
  arc=2pt,
  left=8pt,right=8pt,
  top=6pt,bottom=6pt,
  attach boxed title to top center={yshift=-2mm},
  boxed title style={
    colback=white,
    colframe=black,
    boxrule=0.8pt,
    sharp corners
  },
  #1
}

\usepackage{color-edits}
\addauthor{ak}{Orange}
\addauthor{nv}{ForestGreen}
\addauthor{ah}{gray}

\newcommand{\e}[1]{\texttt{e#1}}

\usepackage[utf8]{inputenc} 
\usepackage[T1]{fontenc}    
\usepackage{url}            
\usepackage{booktabs}       
\usepackage{amsfonts}       
\usepackage{nicefrac}       
\usepackage{microtype}      
\usepackage{natbib}
\usepackage{makecell}
\usepackage{enumitem}
\usepackage{breakcites}
\usepackage{mathrsfs}

\usepackage{algorithm}
\usepackage{verbatim}

\makeatletter
\let\OldStatex\Statex
\renewcommand{\Statex}[1][3]{%
  \setlength\@tempdima{\algorithmicindent}%
  \OldStatex\hskip\dimexpr#1\@tempdima\relax}
\makeatother

\usepackage{multicol}
\usepackage{colortbl}
\usepackage{setspace}
\usepackage{transparent}
\usepackage{upgreek}

\usepackage{inconsolata}
\usepackage[scaled=.90]{helvet}
\usepackage{xspace}

\usepackage{graphicx}

\let\oldparagraph\paragraph

\renewcommand{\paragraph}[1]{\oldparagraph{#1.}}

\title{\huge Select and Improve: Understanding the \\ Mechanics of Post-Training for Reasoning}

\arxiv{
  \author{}
  \date{}
}

\begin{document}

\maketitle

\arxiv{ 
\vspace{-4em}
\begin{center}
\large
\setlength{\tabcolsep}{10pt}
\begin{tabular}{cccc}
\makecell{Akshay Krishnamurthy\footnotemark[1]}
&
\makecell{Audrey Huang\footnotemark[2]}
&
\makecell{Nived Rajaraman\footnotemark[1]}
&
\end{tabular}
\end{center}
\vspace{2em}

\footnotetext[1]{Microsoft Research NYC, \texttt{akshaykr@microsoft.com}, \texttt{nrajaraman@microsoft.com}}
\footnotetext[2]{UIUC, \texttt{audreyh5@illinois.edu}}
}

\begin{abstract}
Reinforcement learning has rapidly emerged as a key component in the training of reasoning and coding models, yet it remains poorly understood from a mechanistic perspective.
We study how and through what underlying processes capabilities are acquired or enhanced via reinforcement learning post-training.
Our analysis, based on controlled math reasoning experiments with \textsc{Qwen-2.5-1.5B}, reveals two core mechanisms: strategy selection and strategy improvement.
Our results highlight the role of SFT data and reinforcement learning data in activating these mechanisms, in particular showing how supervising the model on diverse reasoning strategies can enable strategy selection and how increasing difficulty in reinforcement learning data can enable strategy improvement.
Taken together, our results provide mechanistic insight into RL training and suggest practical interventions to continue scaling reasoning capabilities.
\end{abstract}

\section{Introduction}

Recent progress in advanced reasoning and coding models is largely driven by reinforcement learning (RL) post-training, where a pre-trained base model is further improved through interaction with math, programming, or software engineering problems. It is well documented that this interactive training paradigm leads to significant improvements in model capabilities and state-of-the-art performance. Correspondingly, a rapidly growing body of work has proposed algorithmic and data-centric interventions to the reinforcement learning pipeline, with promising experimental results on academic and industry benchmarks. 

Despite this progress, much of the recent research focuses on improving the RL recipe through interventions. By contrast, comparatively little is understood about the mechanisms by which RL improves the capabilities of the base model. Thus, we ask two central questions:

\begin{quote}
\begin{center}
    \textit{Through what mechanisms does RL lead to improved model capabilities?\\ What is the role of data in enabling these mechanisms?} 
\end{center}
\end{quote}

We answer these questions via a mechanistic study of RL post-training, focusing on math reasoning. Our experiments reveal two core mechanisms: \textbf{strategy selection} which routes problems to existing reasoning patterns learned during pre-training and \textbf{strategy improvement} which improves existing reasoning patterns. We also find that whether these mechanisms activate during training depends strongly on both the pre-training and reinforcement learning datasets. Notably, strategy selection---the largest driver of performance in our experiments---requires that the pre-training data contains reasoning patterns to select among; we do not observe novel patterns emerging through reinforcement learning. 
Meanwhile, strategy improvement requires that the RL dataset consists of more difficult questions than those seen during SFT; RL training with similar problem difficulty does not improve OOD generalization. 

Our experimental setup also elicits two previously-observed phenomena, \textbf{strategy amplification}, where particular reasoning patterns appear more frequently after RL, and \textbf{strategy composition}, where, through RL training, the model improves its ability to chain together multiple complex reasoning steps. However, we find that strategy amplification is an observable effect of strategy selection, and strategy composition is a particular instance of strategy improvement. Thus, we do not view these as separate mechanisms but rather as phenomena that can emerge from the core mechanisms of selection and improvement.

Our findings offer a new perspective into the behavior of RL post-training. We find that RL in the language model setting largely operates by refining reasoning patterns acquired during pre-training. Such refinements can significantly improve performance, but do not induce entirely new behaviors. Thus, our results emphasize the role of high-quality pre-training data as an essential component in scaling reinforcement learning. We hope our findings inspire a more holistic approach to model training and provide a mechanistic foundation for future advances in RL post-training.

\subsection{Related Work}
Reinforcement learning is now a central component in modern language model training pipelines, and a growing body of work studies its behavior, identifying recurring phenomena such as policy sharpening~\citep{yue2025does,wu2025invisible}, entropy reduction, dependence on the pre-trained model, and sensitivity to reward and data designs~\citep{guo2025deepseek,liu2025prorl,shao2025spurious,liu2025understanding}. Our results build on this line of work by introducing a clear taxonomy of mechanisms, and a single experimental setup that manifests several mechanisms, unifying and clarifying prior observations.

Among the specific mechanisms identified in our work, strategy selection was observed to some extent in prior work~\citep{zhao2025echo,tsilivis2025reinforcement,matsutani2025rl,chen2025reshaping}. In contrast with these works, our findings go beyond convergence to a single strategy: we find that RL can \emph{route} problems to different strategies observed in the pretraining data, thereby emphasizing the role of \textit{data diversity} in pretraining.

Strategy amplification has been prominently observed through the emergence of backtracking in DeepSeek-R1~\citep{guo2025deepseek}, and has been studied through the lens of distributional sharpening~\citep{huang2024self,yue2025does,wu2025invisible}. Complementary analyses show that amplification can occur through entropy reduction~\citep{cui2025entropy}, high-entropy branching tokens~\citep{wang2025beyond}, and implicit biases in trajectory ranking~\citep{he2025rewarding}. Distributional amplification can emerge even under spurious rewards, which triggers amplification of pretraining behaviors~\citep{shao2025spurious}. Our experiments associate amplification with selection, offering a mechanistic explanation. 

Regarding strategy improvement and composition, \citep{zhang2025interplay,yuan2025f} demonstrate extrapolative generalization and composition in controlled settings that are closely related to strategy improvement. We observe similar findings and further show that strategy improvement can sometimes be a secondary driver of RL performance relative to strategy selection.

We survey and discuss other work related to these aspects in~\cref{app:related_work}.

\section{Setup}

We consider a standard language model post-training setup involving both supervised fine-tuning (SFT) and reinforcement learning (RL). We adopt the following reinforcement learning notation: prompts $\bx$ and responses $\by$ belong to prompt and response spaces $\cX,\cY$ respectively and a language model $\pi: \cX \to \Delta(\cY)$ associates a distribution over responses to any prompt. Supervised fine-tuning takes a dataset $\Dsft := \{(x_i,y_i)_i\}$ of prompt-response pairs and maximizes\footnote{Our experiments operate in a chain-of-thought setting where both prompts and responses are sequence of tokens.~\cref{eq:sft_obj} corresponds to standard token-level cross entropy loss minimization in this case.}
\begin{align}
\Lsft(\pi; \Dsft) := \sum_{(x,y) \in \Dsft} \log \pi(y \mid x)\label{eq:sft_obj}
\end{align}
using gradient-based optimization, initialized from a pre-trained language model $\pibase$. In RL, we have a dataset $\Drl := \{(x_i)_i\}$ of prompts and a reward function $R:\cX\times\cY \to \{ 0,1 \}$ determining correctness (or quality) of response $y$ to prompt $x$. Using these, we seek to maximize
\begin{align}
    \Lrl(\pi;\Drl) := \sum_{x \in \Drl} \bbE_{y \sim \pi(\cdot \mid{} x)}[R(x,y)],\label{eq:rl_obj}
\end{align}
again using gradient-based optimization, initialized now from the policy $\pisft$ obtained from the SFT stage. Models are primarily evaluated via the reinforcement learning objective~\cref{eq:rl_obj} on a held out dataset of prompts $\Dval := \{(x_i)_i\}$.

\subsection{Experimental setup: Finite-field arithmetic task}


Throughout our experiments, we focus on a synthetic finite-field arithmetic task, which captures a surprising depth of behaviors through subtle changes to the basic version of the task and the training setup. Following the protocol above, we always SFT a base model (\textsc{Qwen2.5-1.5B-Instruct}) on some SFT dataset, and continue training via RL on some RL dataset, which may be from a different distribution. First we describe the task.

\paragraph{Prompts} 
Towards building a mechanistic understanding, it is essential to carefully control the set of reasoning strategies available to the model before RL training. At the same time, we want to operate at a reasonable scale so must rely on open source pre-trained models. To break away from the math reasoning ability contained within such pre-trained base models, we construct a finite-field arithmetic task where integers are replaced by abstract symbols, and the goal is to manipulate them to compute the answer. The problem begins with a randomly generated permutation to define the elements of the finite field. The subsequent computational problem takes one of two forms: evaluation or inversion (see~\cref{fig:problems}).
\begin{itemize}[leftmargin=1.25em]
    \item \textbf{Evaluation problems.} Evaluation problems are those where the model is given a sequence of arithmetic operations, and is tasked with performing them in order.
    \item \textbf{Inversion problems.} Inversion problems, on the other hand, correspond to solving for an unknown variable. The model is given an unknown input $x$, and a sequence of arithmetic operations which manipulate $x$ in various ways resulting in a numerical value. The model must solve for the value of $x$. 
\end{itemize}

\noindent

\begin{figure}[t]
\input{figs/dataset_fig}
\caption{Examples of evaluation and inversion problems and forward and backward solutions over GF(11).}
\label{fig:problems}
\end{figure}

For both evaluation and inversion problems, the difficulty of the task is captured by the number of steps of arithmetic manipulation required to compute the final answer.
For the main experiments, we operate over GF(11) as shown in~\cref{fig:problems} and we provide some results with GF(13) in~\cref{app:exp_results} for robustness.
We always SFT under the uniform distribution over evaluation and inversion problems with number of arithmetic steps chosen uniformly from 2-5. 

\paragraph{SFT data}
We next describe the responses $y$ utilized in the SFT data. These comprise the main reasoning strategies we embed into the model, which can be applied regardless of the problem type (evaluation or inversion), and can be thought of as different ways to solve the same problem. We focus on two broad strategies, which we refer to as \emph{forward} and \emph{backward} reasoning (See~\cref{fig:problems}).

\begin{itemize}[leftmargin=1.25em]
    \item \textbf{Forward reasoning.} In forward reasoning, the model's chain of thought proceeds by starting from the initial input and performing the arithmetic operations in order, until it reaches the final answer. Forward reasoning is the natural strategy for evaluation problems, where the model is given an initial input and a sequence of operations to perform. The model can simply read off the operations and apply them in order to compute the final answer. For inversion problems, forward reasoning is less natural, as the model must first read off the operations and then figure out how to undo them in order to solve for the unknown input.
    \item \textbf{Backward reasoning.} In backward reasoning, the model's chain of thought proceeds by starting from the final answer and working backwards through the arithmetic operations to determine the initial input. Backward reasoning is the natural strategy for inversion problems, as it can be viewed as applying forward reasoning on an associated evaluation problem (that of applying inverse operations on the final output).
\end{itemize}

\paragraph{RL data and rewards} To better understand the interplay between SFT and RL data, we perform RL training under a variety of distribution shifts:
\begin{itemize}[leftmargin=1.25em]
    \item \textbf{Extended difficulty.} In our main setting, the prompts in the RL dataset have 6-9 arithmetic steps (chosen uniformly), reflecting more challenging reasoning problems. We conduct further experiments where problems have 6-15 arithmetic steps, representing an extreme distribution shift in difficulty.
    \item \textbf{Tilting.} In our main setting, we maintain a uniform distribution over evaluation and inversion problems, reflecting no tilting relative to the SFT dataset. In subsequent experiments, we tilt toward evaluation problems at 75-25 and 90-10 proportions. 
\end{itemize}
For rewards, we always consider outcome level reward, that is, whether the final answer reported by the model is correct for the given arithmetic question. 

\subsection{Training } \label{sec:training}

We take \textsc{Qwen2.5-1.5B-Instruct} as the base pre-trained model. For training, we use LoRA with rank 64, $\alpha=128$, and dropout 0.05. For SFT, we use a dataset of 2048 prompt-response pairs, and train for 20 epochs with AdamW with learning rate $2\times 10^{-4}$, batch size 16, and warmup-stable learning rate schedule with linear warmup for 1 epoch. For RL, we use datasets of 1024 prompts and always use GRPO with outcome rewards, batch size 32, group size 8, KL parameter $\beta=0.05$, and temperature 0.7 for rollout generation. We train with AdamW with learning rate $1\times 10^{-5}$ for 500-1000 steps, saving checkpoints every 25 steps. For each experiment, we use 3 seeds for SFT and a single RL run starting from each SFT checkpoint, for 3 total RL runs. In~\cref{app:exp_results} we demonstrate via ablations that our results are robust to various modifications to hyperparameters. 

\section{Experimental results}
In this section, we present our experimental results, identifying strategy selection and strategy improvement as the two central mechanisms by which RL improves reasoning performance. 

\begin{figure}
    \centering
    \includegraphics[width=1\linewidth]{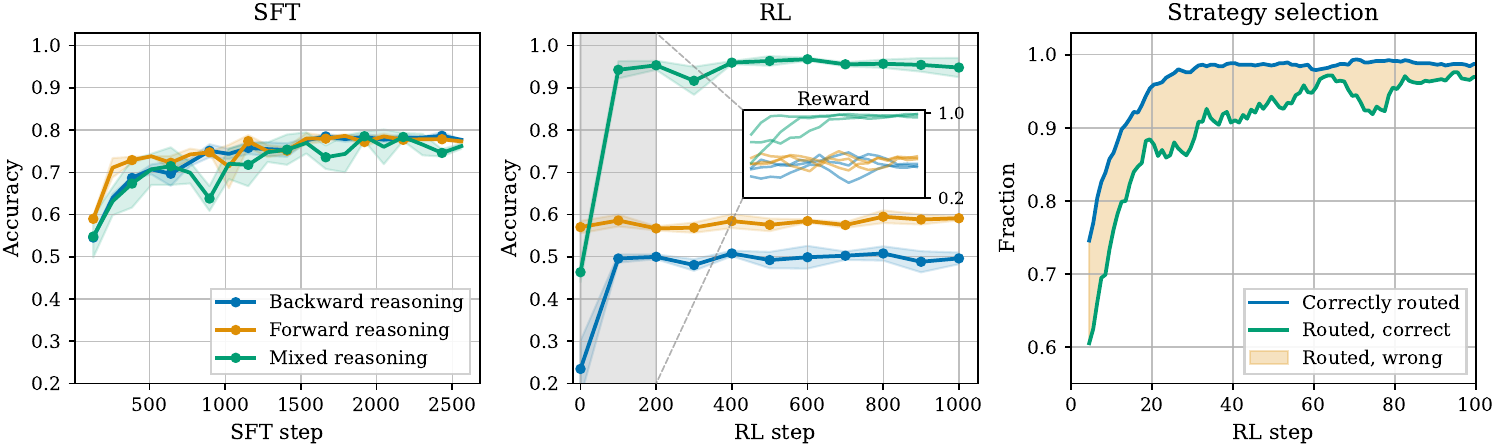}
    \caption{Main results in $\textrm{GF}(11)$ setting, demonstrating both strategy selection and improvement. Left panel shows accuracy on held-out 2-5 step problems over the course of SFT training. Center panel shows accuracy on held-out 6-9 step problems over the course of RL training, with inset showing per step (in-sample) reward curves. Right panel focuses on the learning dynamics of FB models. A simple rule-based classifier identifying strategy type shows that performance improvements are largely driven by \emph{strategy selection} with \emph{strategy improvement} contributing to a lesser degree.}
    \label{fig:main}
\end{figure}

\subsection{Mechanism 1: Strategy Selection}

\begin{figure}[t]
    \centering
    \includegraphics[width=1\linewidth]{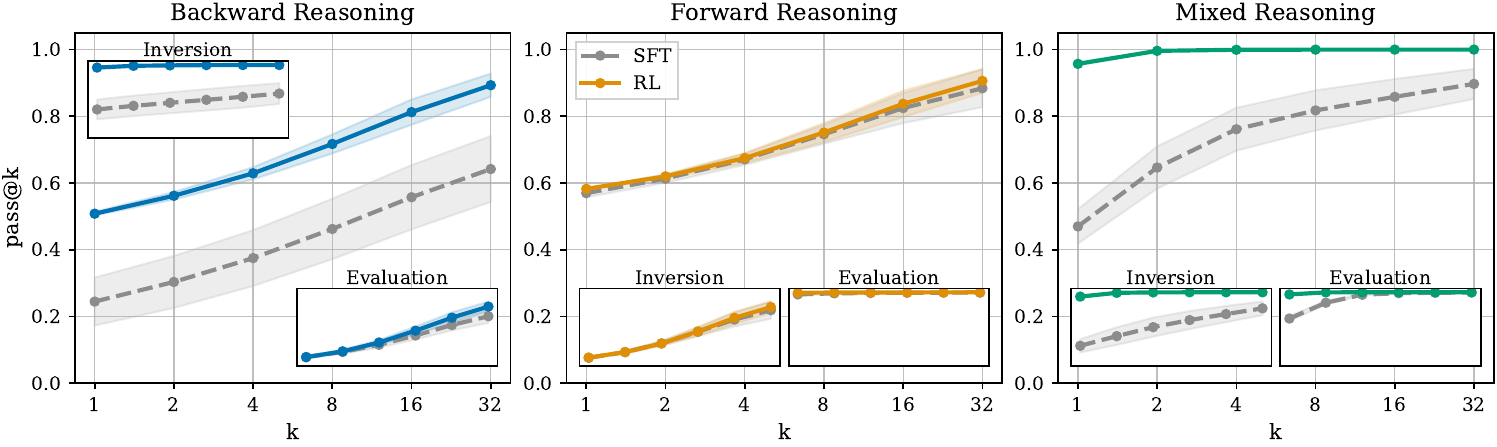}
    \caption{Pass@k results for the best SFT and RL checkpoints in the main $\textrm{GF}(11)$ setting. Main panels visualize aggregated results for Backward (left), Forward (center) and mixed (right) reasoning, with insets displaying results for inversion and evaluation problem subsets. Backward reasoning, particularly on inversion problems, improves substantially through RL. Forward reasoning on evaluation problems is already saturated despite increased problem difficulty. Backward on evaluation (resp. forward on inversion) show no improvement through RL. Finally mixed reasoning improves on both subsets and achieves near-perfect pass@1 performance in aggregate.}
    \label{fig:passk_main}
\end{figure}

\begin{figure}[t]
    \centering
    \includegraphics[width=1\linewidth]{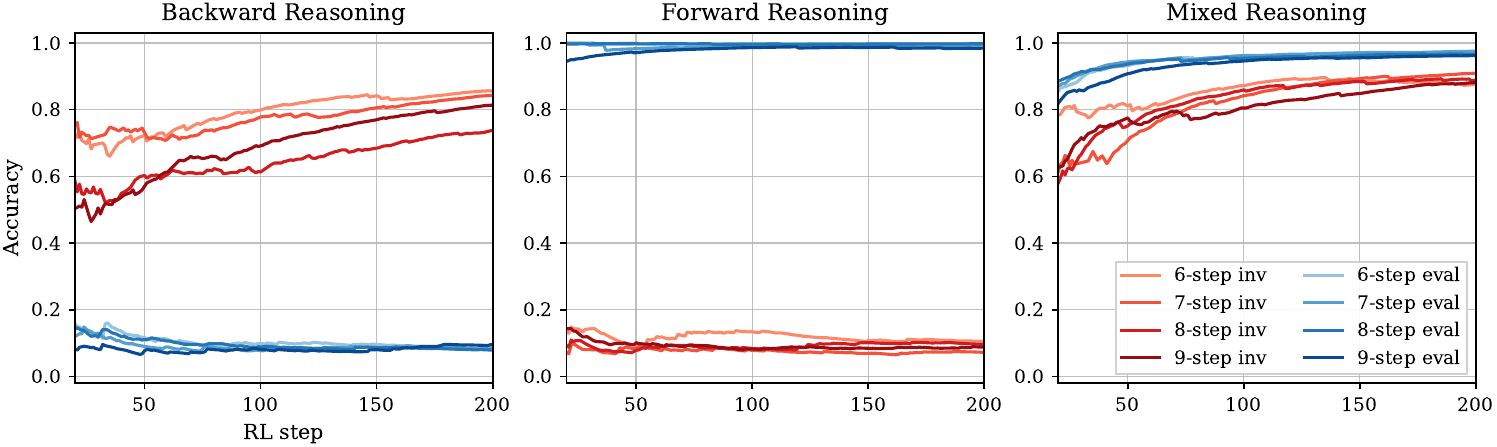}
    \caption{Breakdown of RL training dynamics in the main $\textrm{GF}(11)$ setting. We visualize the running average reward for each problem type (inversion/evaluation) and each number of steps (6-9). Backward reasoning on evaluation problems (resp. forward reasoning on inversion problems) shows no improvement with RL training. For mixed reasoning, we see improvements in all problem subgroups.}
    \label{fig:breakdown_main}
\end{figure}

The first experiment identifies both strategy selection and strategy improvement mechanisms. In this experiment, we generate three SFT datasets by adjusting the responses to contain $(a)$ forward-only reasoning (F), $(b)$ backward-only reasoning (B), and $(c)$ mixed forward+backward (FB) reasoning, where in the latter each problem is solved with one of the two strategies chosen uniformly. 
We train SFT variants on each of these three datasets, and then perform RL training on harder 6-9 step problems over $\text{GF}(11)$ and $\text{GF}(13)$. Results are visualized in~\Cref{fig:main,fig:passk_main,fig:breakdown_main}. 


Recall that forward reasoning is the natural strategy for evaluation problems (direct step-by-step computation), while backward reasoning is natural for inversion problems (undoing operations from the known result). This design is reflected in the experimental results, where each single-strategy model masters its natural problem type but struggles on the other: the F-only model achieves ${\sim}100\%$ on 2-5 step evaluation but only ${\sim}60$-$65\%$ on inversion, while the B-only model achieves ${\sim}100\%$ on inversion but only ${\sim}55$-$60\%$ on evaluation. Both models achieve ${\sim}80\%$ accuracy in aggregate. 
The FB model, also achieves ${\sim}80\%$ accuracy in aggregate (left panel of ~\cref{fig:main}) but for a different reason: it selects strategies at random for each problem and has a 50\% chance of picking the ``wrong'' strategy. 


However, the emergent behavior after RL training is very different. 
For the F-only (resp. B-only) model, RL learns to master the forward (resp. backward) strategy that it has already seen in the SFT data, enabling the accuracy to improve on evaluation (resp. inversion) problems. The other problem type sees essentially no improvement in performance over RL training. Overall, for the F-/B-models, RL learns to transfer the ${\sim}80\%$ SFT performance on $2$-$5$ step problems to the harder $6$-$9$ step problems, but the accuracy plateaus at ${\sim}60\%$ for F-models and ${\sim}50\%$ for B-models thereafter.\footnote{Examining the left and center panels of~\cref{fig:passk_main}, this drop in performance arises because backward (resp. forward) reasoning is worse on the harder evaluation (resp. inversion) problems than on the easier ones. Further, backward reasoning is slightly harder to learn in general than forward reasoning.}
Without access to an alternative reasoning strategy, there is no better strategy for RL to select.

In contrast, the FB model tells a strikingly different story. Despite starting from an SFT checkpoint with comparable performance, RL rapidly drives accuracy to ${\geq}95\%$ on both problem types across all seeds. This is depicted for $\text{GF}(11)$ in the center panel of~\cref{fig:main}. 
The right panel shows that the mechanism behind this is \emph{strategy selection}: 
the SFT mixture exposes the model to both forward and backward reasoning patterns, and RL rapidly learns to dispatch each problem to the appropriate strategy---forward for evaluation, backward for inversion. This routing is the dominant driver of improvement. The FB model succeeds not because it learns a new reasoning strategy during RL, but because the SFT data endowed it with a richer repertoire of strategies to choose from.

\subsubsection{Case Study: Strategy Amplification}

\begin{figure}
    \centering
    \includegraphics[width=1\linewidth]{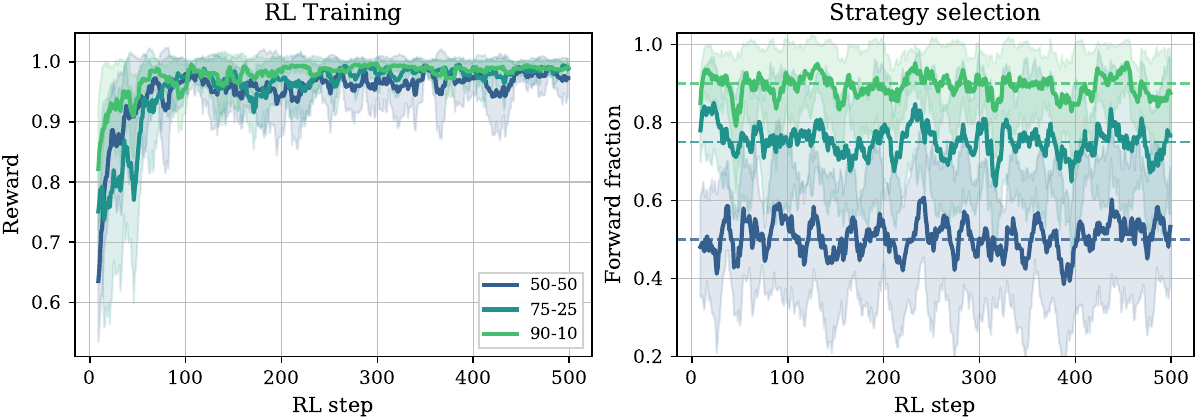}
    \caption{Results for mixed reasoning models in the skewed setting with GF(11). Left: The (smoothed) reward under three RL problem distributions reflecting skew toward evaluation problems. Right: the fraction of forward generations over time precisely matches the fraction of evaluation problems in the dataset.}
    \label{fig:skewed_main}
    \vspace{-1em}
\end{figure}

Strategy amplification occurs when a reasoning pattern appears much more frequently after reinforcement learning than before. To study strategy amplification in our setting, we focus on FB models on RL datasets that over-represent evaluation problems (i.e., tilted datasets). In~\cref{fig:skewed_main}, we find that in three different RL dataset compositions (50-50, 75-25, and 90-10 evaluation-inversion proportions), RL post-training an FB model yields near-perfect accuracy. In all cases, analysis of the roll-outs shows that the fraction of forward-style reasoning closely matches the fraction of evaluation problems in the dataset, reflecting strategy amplification in the tilted settings. However, the underlying mechanism is strategy selection; RL training routes problems to the best reasoning pattern, so a skew in problems results in a skew in observable reasoning patterns. Strategy amplification is an observable effect of strategy routing. 

This finding may shed light on prior works identifying amplification of reasoning patterns such as backtracking and the so-called ``aha moments''~\citep{guo2025deepseek}. Consistent with our results, these prior works show that RL can amplify reasoning patterns obtained through pre-training and SFT. Our controlled experiments identify strategy selection as a potential cause for amplification: the process of identifying useful strategies for the RL dataset leads to increases in observable features such as backtracking. 

\begin{tcolorbox}[
  enhanced,
  colback=gray!5,
  colframe=gray!50!black,
  arc=4pt,
  boxrule=0.6pt,
  left=6pt, right=6pt, top=6pt, bottom=6pt,
  title={\textbf{Strategy selection takeaways}},
  fonttitle=\small,
  coltitle=white,
  colbacktitle=gray!50!black,
]
\begin{itemize}[leftmargin=1.25em]
    \item Strategy selection occurs when RL associates problems to reasoning patterns.
    \item Strategy selection leads to rapid performance improvements during RL.
    \item Strategy selection is an underlying cause of strategy amplification.
    \item Strategy selection relies on diverse reasoning in pre-training and SFT data. 
\end{itemize}
\end{tcolorbox}

\subsection{Mechanism 2: Strategy Improvement}

Strategy improvement occurs when the basic reasoning patterns learned during pretraining/SFT are enhanced during RL training. In our setting, we train on problems of increased difficulty to probe strategy improvement. Indeed, this mechanism can already be seen in the main experiments visualized in~\cref{fig:main,fig:passk_main,fig:breakdown_main}. Specifically, in the right panel of~\cref{fig:main} we see that the fraction of correctly routed but incorrect solution attempts decreases over training, reflecting improvement in the underlying reasoning pattern. This is corroborated by the RL training dynamics visualized in~\cref{fig:breakdown_main} where disaggregated performance---broken down into segments of question type and difficulty---is monotonically improving for all segments. Moreover, comparing the time scales of strategy selection (40 steps) and disaggregated reward, we see substantial performance improvements occur well after selection is completed. Finally, focusing on the B-only models, where strategy selection is suppressed, RL improves to perfect accuracy on inversion problems (upper inset of left panel of~\cref{fig:passk_main}).  

\begin{table}
    \centering
\begin{tabular}{lccccccccc}
\toprule
& \multicolumn{3}{c}{\textbf{Backward}} & \multicolumn{3}{c}{\textbf{Forward}} & \multicolumn{3}{c}{\textbf{Mixed}} \\
\cmidrule(lr){2-4} \cmidrule(lr){5-7} \cmidrule(lr){8-10}
\textbf{Training} & All & Inv & Eval & All & Inv & Eval & All & Inv & Eval \\
\midrule
SFT        & 24.5 & 38.9 & 11.9 & 57.0 & 10.7 & 97.1 & 47.0 & 27.1 & 64.3 \\
RL (2--5)  & 18.4 & 25.6 & 10.1 & 48.0 &  9.3 & 91.7 & 60.0 & 32.2 & 91.5 \\
RL (6--9)  & \textbf{50.8} & \textbf{96.0} & 11.6 & 58.3 & 11.0 & \textbf{99.3} & \textbf{95.7} & \textbf{94.1} & \textbf{97.1} \\
\bottomrule
\end{tabular}\vspace{0.5em}
\caption{Pass@1 accuracy on 6--9 step problems for different models. SFT models trained on 2--5 step data, models trained via RL on both 2--5 step and 6--9 step problems. For forward and backward reasoning, RL on 2--5 step problems (in distribution) degrades performance slightly, while RL on 6-9 step performance induces strategy improvement (especially for backward models). For mixed reasoning, in-distribution RL only induces strategy selection, while RL on 6--9 step problems induces both strategy selection and improvement. }
\label{tab:easy-vs-hard-rl}
\vspace{-1em}
\end{table}


We validate the robustness of these results by considering the even harder RL dataset comprising problems with 6-15 arithmetic steps. As the findings are similar, we defer the details to the appendix.

In~\cref{tab:easy-vs-hard-rl} we study the role of the RL dataset in eliciting strategy improvement. Initializing from the same SFT checkpoints, we perform RL on an \emph{easy} dataset consisting of 2-5 arithmetic step problems, which matches the distribution of the SFT data. We evaluate the SFT checkpoints, these easy post-RL models, and the models that were RL-trained on 6--9 step problems, on a validation set of 6--9 step problems and record the pass@1 performance in aggregate and disaggregated into problem type. We see that RL training on easy data does not lead to performance improvements for Backward- and Forward-only models, where strategy selection is suppressed; no strategy improvement occurs in these settings. For mixed reasoning, RL on easy data does lead to performance improvements, but this can be attributed to strategy selection; strategy improvement only appears when RL training on the harder dataset, where it yields further performance improvements. 

\subsubsection{Case Study: Strategy Improvement for Composition}

We consider an alternate setting demonstrating strategy improvement through skill composition. Here, we introduce \emph{composition problems}, where a prompt chains together $n$ evaluation and inversion sub-problems over the same finite field. The model must solve each sub-problem sequentially and propagate intermediate results across the full chain, testing a qualitatively different capability from solving individual problems. Examples of such problems are described in the appendix in~\cref{fig:composition_example}.

\begin{figure}
    \centering
    \includegraphics[width=0.85\linewidth]{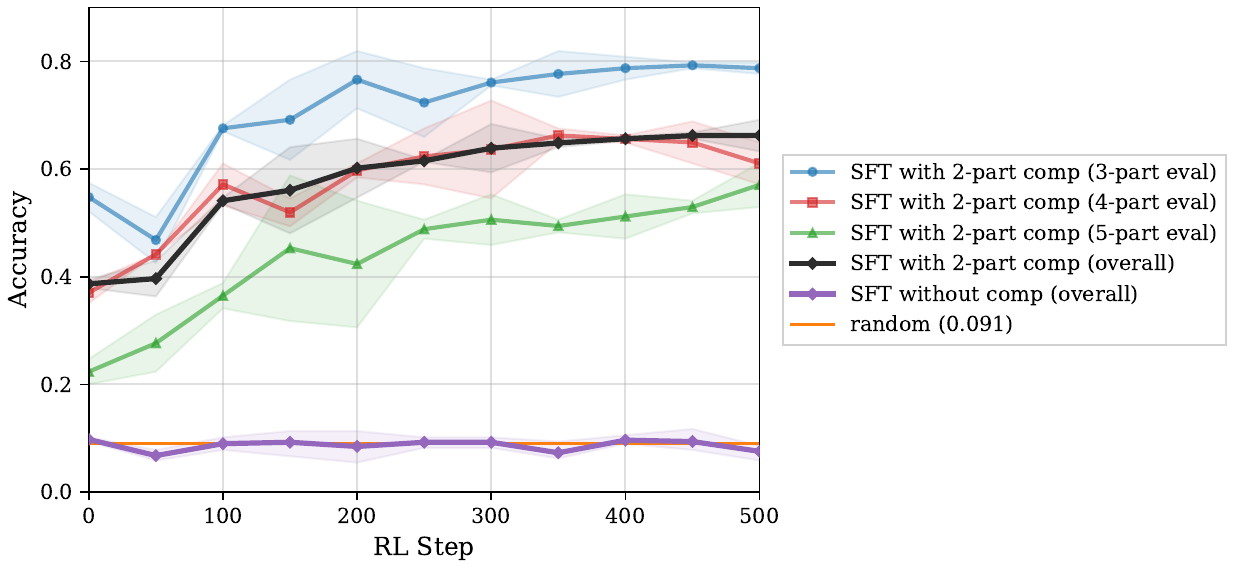}
    \caption{Main results in the $\textrm{GF}(11)$ setting with composition, demonstrating strategy improvement. The {\color{red!75!gray} \bfseries red}, {\color{green!50!gray} \bfseries green} and {\color{blue!65!green!70!gray} \bfseries blue} curves show accuracy on held-out 3-5 part composition problems over the course of RL training when the SFT mix contains 2-part composition problems. The {\color{red!50!blue} \bfseries purple} curve tracks the average evaluation performance on 3-5 part problems over the course of RL training, when the SFT dataset only contains inversion and evaluation problems.}
    \label{fig:composition}
    \vspace{-1em}
\end{figure}

We compare two SFT configurations: one including 2-part composition problems (each part is a $1$-step evaluation or inversion problem) alongside standard length 2-5 evaluation and inversion problems, and the other with only length 2-5 evaluation and inversion problems. In both cases, we perform RL on harder 3-5 part composition problems (each part is a $1$- or $2$-step inversion or evaluation sub-problem). Results are visualized in~\cref{fig:composition}.

When 2-part composition is present in the SFT data (left panel of~\cref{fig:composition}), RL yields substantial improvements on 3-5 part problems, where each part may also have more steps. Thus RL extends the composition skill seeded by the 2-part SFT examples to longer chains. In contrast, without any composition data in the SFT mix (right panel), RL yields no improvement, despite the model possessing the individual sub-skills. This illustrates strategy improvement for composition: RL strengthens and generalizes the ability to compose to harder instances, but does not learn composition from scratch.

Our results are in tension with~\citep{yuan2025f} who show that RL can teach the model to compose atomic skills when they are already present in the model. This phenomenon appears to be nuanced: in our setting, some degree of compositional data in the SFT dataset appears essential for RL to improve compositional reasoning.

\begin{tcolorbox}[
  enhanced,
  colback=gray!5,
  colframe=gray!50!black,
  arc=4pt,
  boxrule=0.6pt,
  left=6pt, right=6pt, top=6pt, bottom=6pt,
  title={\textbf{Strategy improvement takeaways}},
  fonttitle=\small,
  coltitle=white,
  colbacktitle=gray!50!black,
]
\begin{itemize}[leftmargin=1.25em]
    \item Strategy improvement occurs when RL improves existing reasoning patterns.
    \item Strategy improvement operates on a slower timescale than selection.
    \item Composition benefits from strategy improvement, but RL does not learn composition from scratch.
    \item Strategy improvement relies on increasing difficulty in RL data. 
\end{itemize}
\end{tcolorbox}




\section{Discussion}

In this paper, we present a study of reinforcement learning for math reasoning through controlled experiments on a finite-field arithmetic task, and identify two core mechanisms by which RL improves performance: strategy selection and strategy improvement. The emergence of these mechanisms depends strongly on the pre-training / SFT datasets and what kinds of solution strategies they expose to the model.

When SFT data exposes the model to multiple reasoning strategies, RL rapidly learns to route problems to the most effective one, achieving near-perfect accuracy. In our setting, models trained with a single strategy are fundamentally limited and RL cannot compensate. This underscores that efforts to scale RL-based reasoning should prioritize diversity in pre-training and SFT data, not only the downstream RL recipe. Strategy improvement is essential for generalization to harder problems, but requires the RL dataset to present problems of greater difficulty than those seen during SFT; training on problems of comparable difficulty yields no improvement. This suggests that RL datasets should be designed to contain a difficulty curriculum.

While our experiments do not identify settings where RL induces novel reasoning capabilities, a phenomenon that appears to emerge from continued RL on challenging datasets~\citep{liu2025prorl}, understanding the mechanisms through which RL moves beyond selection and refinement of pre-existing strategies is perhaps the most important open question raised by our work. More broadly, our results suggest that viewing RL post-training in isolation is insufficient. The effectiveness of RL is tightly coupled with the quality and diversity of supervision in the pre-RL stages, which determine the repertoire of strategies available for selection and improvement. We hope that this study encourages more integrated approaches to the full training pipeline, and provides a foundation for principled interventions that continue to scale reasoning capabilities.



\bibliographystyle{alpha}
\bibliography{refs}


\clearpage

\appendix

\appendix

\section{Extended Related Work}
\label{app:related_work}

\paragraph{RL for language models}
The rapid adoption of RL in language model pipelines has spurred a large body of empirical and algorithmic work on training recipes, data curation, and scalable implementations~\citep{guo2025deepseek,liu2025prorl,liu2025understanding}. Alongside these efforts, several studies have attempted to characterize recurring phenomena in RL-trained models: policy distributions tend to sharpen over training~\citep{yue2025does,wu2025invisible}, entropy decreases monotonically, and final performance depends heavily on the quality of the pre-trained checkpoint and the design of the reward signal. Notably,~\citep{liu2025understanding} critically examine R1-Zero-like training (RL directly on base models without SFT) and find that behaviors such as ``Aha moments'' are already present in base models prior to RL, and that Qwen2.5 base models exhibit strong reasoning capabilities even without structured prompting, suggesting significant pretraining biases. These observations underscore a recurring theme in the literature, and a central finding of our work: the effectiveness of RL post-training is tightly coupled with what the base model has already learned.

\paragraph{Strategy selection}
Several prior works have observed that RL concentrates probability mass on a subset of reasoning behaviors available in the base model.~\citep{zhao2025echo} shows that, when pre-training data contains multiple output formats, RL consistently converges to a single format, which may not be the most performant. \citep{tsilivis2025reinforcement} study a controlled setting where the model is pretrained with multiple solution strategies, and observe that RL converges to the single most accurate one. Our work differs from these in an important respect: rather than convergence to a single dominant strategy, we observe that RL learns to \emph{route} different problem types to different strategies present in the pretraining data, highlighting that data diversity in pre-training is what enables this richer form of selection.

\paragraph{Strategy improvement}
A number of works demonstrate that RL can yield sustained performance gains, particularly with well-designed data mixtures~\citep{liu2025prorl,sun2025omega,sun2025rl}, though the mechanisms driving these gains are often left unexamined. In controlled settings,~\citep{zhang2025interplay} identify what they term extrapolative generalization: when RL data difficulty slightly exceeds that of pre-training, the model's reasoning patterns strengthen and transfer to harder problems.~\citep{yuan2025f} provide a complementary perspective, showing that RL can compose pre-existing atomic skills into new composite capabilities in a synthetic math reasoning task. From a different angle,~\citep{wen2025rlvr} argue that RLVR can genuinely extend the reasoning boundary beyond pure distribution sharpening. Our controlled experiments place them in a broader context: we show that strategy improvement, while important for out-of-distribution generalization, often plays a secondary role relative to strategy selection when both mechanisms are active.

\paragraph{Strategy amplification}
The amplification of certain reasoning behaviors during RL---most visibly backtracking and self-correction---was brought to wide attention by DeepSeek-R1~\citep{guo2025deepseek} and has since been examined from multiple angles.~\citep{huang2024self} introduced the sharpening framework, formalizing how RL concentrates probability mass on successful trajectories already accessible to the base model; subsequent work confirmed this picture empirically~\citep{yue2025does,wu2025invisible}. Our experiments offer a unifying perspective: we show that amplification can be understood as an observable consequence of strategy selection, where the model's learned routing toward effective strategies naturally increases the prevalence of associated reasoning patterns.

\section{Additional Experimental Results}
\label{app:exp_results}

In this section we discuss several additional experimental results. 

\begin{table}
\centering
\caption{SFT learning rate ablation on GF(11). Best validation cross-entropy and corresponding epoch across 20 epochs of training. All learning rates achieve nearly identical performance. $^\star$ denotes the default configuration used in all main experiments.}
\begin{tabular}{lccc}
\toprule
& \textbf{lr=1e-4} & \textbf{lr=2e-4}$^\star$ & \textbf{lr=5e-4} \\
\midrule
B  & 0.0856 (ep 6) & 0.0859 (ep 5) & 0.0863 (ep 6) \\
F  & 0.0857 (ep 6) & 0.0858 (ep 3) & 0.0862 (ep 5) \\
FB & 0.0879 (ep 5) & 0.0879 (ep 6) & 0.0884 (ep 6) \\
\bottomrule
\end{tabular}

\label{tab:sft_ablation}
\end{table}

\begin{table}
\centering
\caption{RL hyperparameter ablation on GF(11). Final reward (smoothed over the last 100 training steps) on 6--9 step problems. $^\star$ denotes the default configuration used in all main experiments. Observe that these choices are near optimal for forward, backward, and mixed reasoning models.}
\begin{tabular}{lcccc}
\toprule
& \multicolumn{2}{c}{\textbf{group\_size=8}} & \multicolumn{2}{c}{\textbf{group\_size=16}} \\
\cmidrule(lr){2-3} \cmidrule(lr){4-5}
& $\beta$=0.05 & $\beta$=0.1 & $\beta$=0.05 & $\beta$=0.1 \\
\midrule
B, lr=1e-5  & 53\%$^\star$ & 53\% & ---  & 58\% \\
B, lr=5e-5  & 14\% & 55\% & 51\% & 53\% \\
\addlinespace
F, lr=1e-5  & 60\%$^\star$ & 58\% & 51\% & 52\% \\
F, lr=5e-5  & 54\% & 56\% & 51\% & 48\% \\
\addlinespace
FB, lr=1e-5 & 97\%$^\star$ & 96\% & 98\% & 97\% \\
FB, lr=5e-5 & 54\% & 96\% & 51\% & 52\% \\
\bottomrule
\end{tabular}

\label{tab:rl_ablation}
\end{table}

\paragraph{Ablations}
First, we ran several ablations to our SFT and RL training recipes to evaluate robustness of our findings. In our setting, training (especially SFT) is substantially less compute-intensive than downstream evaluation, which requires collecting a large number of generations from the model. Consequently, for the ablation study we focus on evaluation metrics that are less compute intensive to collect, specifically validation cross entropy for SFT and online reward for RL. 

Results are displayed in~\cref{tab:sft_ablation} and~\cref{tab:rl_ablation} respectively. For SFT, we consider two alternative learning rates from our default choice of 2$\times 10^{-4}$ and find that results are extremely robust to this choice. As we always select the best SFT checkpoint using generation-based evaluation, the differences in best epoch are not meaningful for our experiments. For RL, we vary learning rate, KL regularization coefficient $\beta$ and the group size used in GRPO. We find that our defaults (learning rate $1\times{}10^{-5}$, $\beta=0.05$ and group size 8) are essentially the best performing choices for all three types of reasoning models. 

\begin{figure}[t]
    \centering
    \includegraphics[width=1\linewidth]{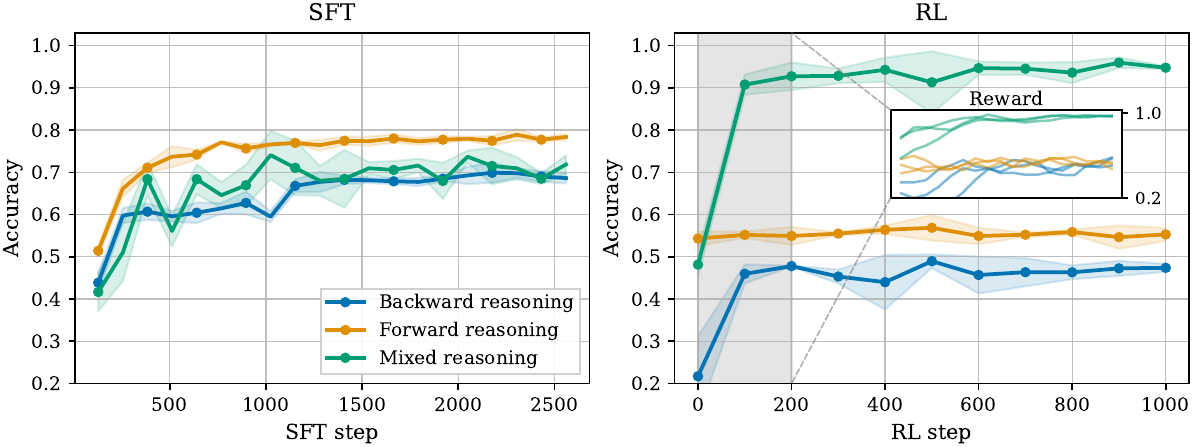}
    \caption{Main results for GF(13). Left: accuracy (measured via outcome correctness on model rollouts) over the course of SFT on the easier (2-5 step) SFT validation dataset, showing that all models saturate at 70-80\% accuracy. Right: accuracy over the course of RL training on the harder (6-9 step) RL validation dataset with inset showing individual run reward curves. As in the main GF(11) setting, mixed reasoning during pre-training enables substantial improvements with RL and this is largely driven by strategy selection.}
    \label{fig:gf13_main}
\end{figure}

\begin{figure}[t]
    \centering
    \includegraphics[width=1\linewidth]{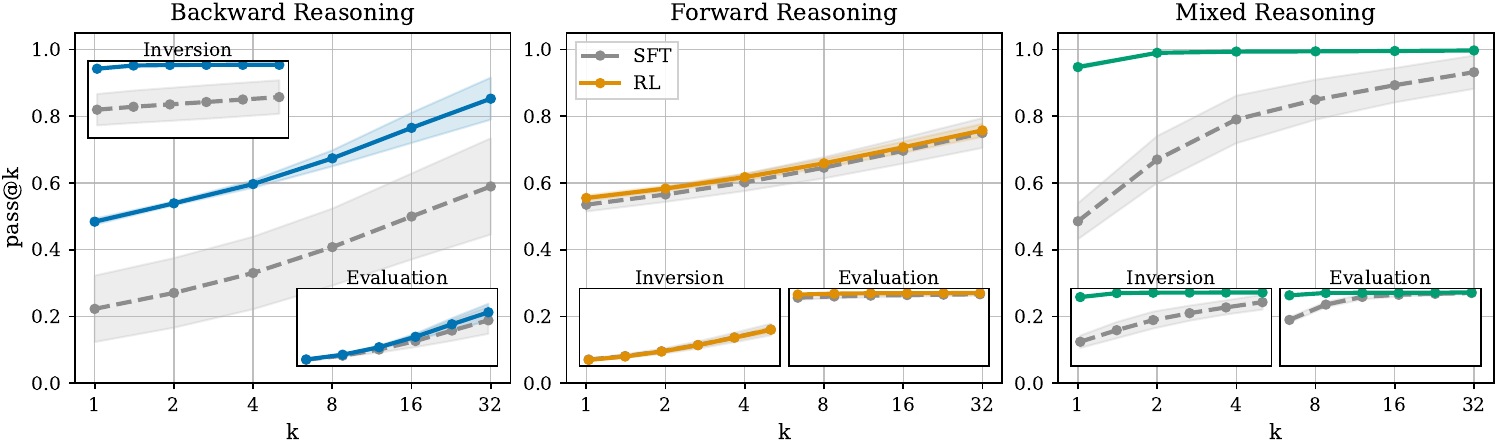}
    \caption{Pass@k results and per-problem-type breakdown for GF(13), paralleling the results in~\cref{fig:passk_main}. Improvements for backward (resp. forward) reasoning models are entirely driven by improvements in inversion (resp. evaluation) problems. Mixed reasoning models achieve near-perfect accuracy on both problem types. }
    \label{fig:gf13_passk}
\end{figure}

\paragraph{Extension to GF(13)} As a further robustness check, we also experiment with arithmetic reasoning over a larger field, namely GF(13). Here we reproduce the main experiments, where we SFT on problems requiring 2-5 arithmetic steps, and RL on problems requiring 6-9 arithmetic steps, where both datasets have a uniform mixture of inversion and evaluation questions. The findings, visualized in~\cref{fig:gf13_main,fig:gf13_passk}, echo those in~\cref{fig:main,fig:passk_main}. Namely, all three models achieve similar performance on the 2-5 step dataset after SFT, but only the mixed reasoning model is able to achieve near-perfect performance on the harder 6-9 step dataset after RL.

\begin{figure}[t]
    \centering
    \includegraphics[width=1\linewidth]{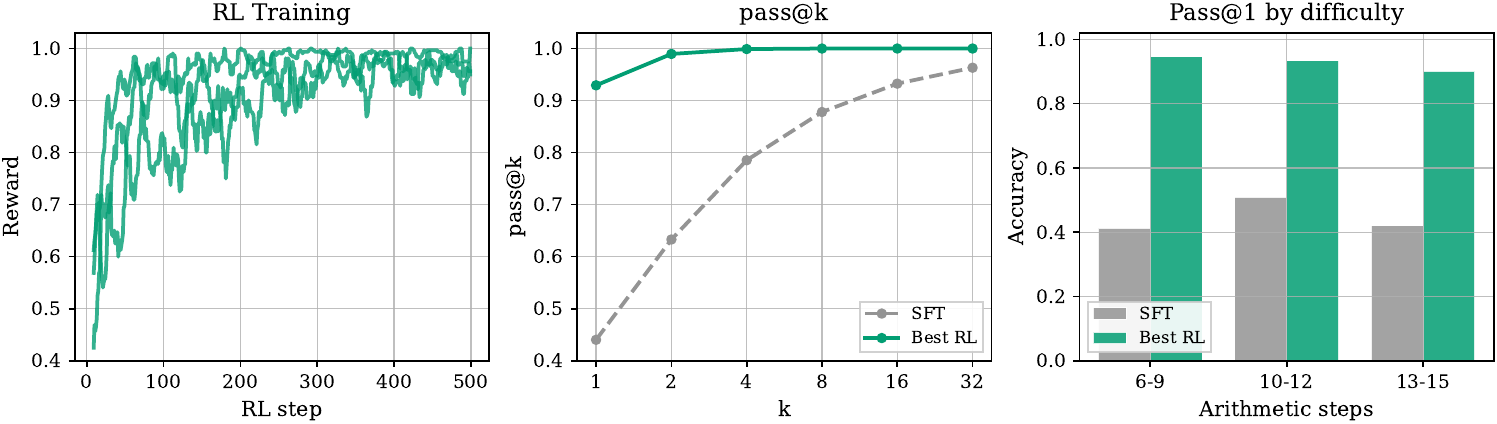}
    \caption{Results for RL training of mixed reasoning models in the extended setting, where the RL dataset contains problems with 6-15 arithmetic steps. Left: reward curves for three independent seeds of RL training. Center: pass@k curves for the SFT initialization and the best RL checkpoint showing substantial performance improvements with RL. Right: pass@1 performance across problem types, stratified by number of arithmetic steps. Consistent with the findings in the main GF(11) experiments, we see that RL rapidly learns to get near-perfect accuracy on this much harder dataset, substantially improving on the SFT initialization.}
    \label{fig:ext_main}
\end{figure}

\begin{figure}[t]
    \centering
    \includegraphics[width=1\linewidth]{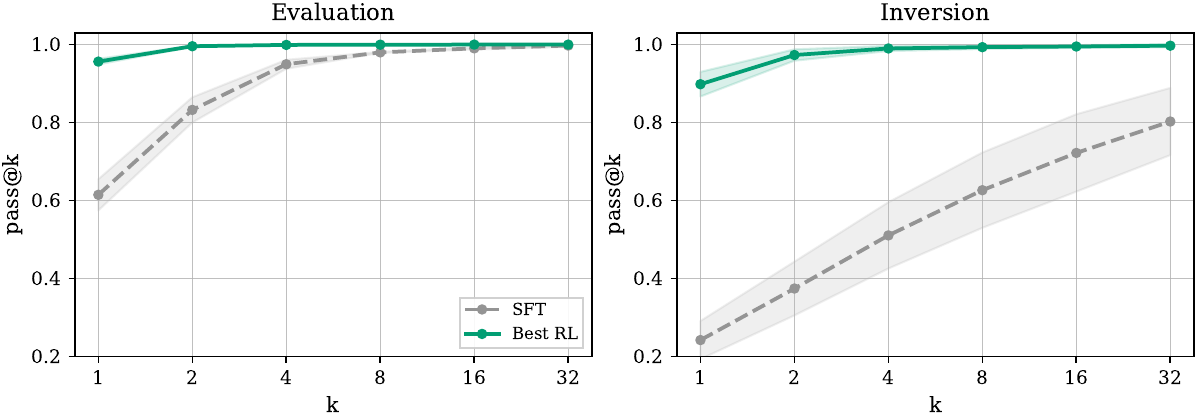}
    \caption{Pass@k results and per-problem-type breakdown for the extended GF(11) setting, where the RL training data consists of problems with 6-15 arithmetic steps. Consistent with findings from the main experiments, RL rapidly achieves near-perfect accuracy, in particular eliciting substantial performance gains on inversion problems.}
    \label{fig:ext_breakdown}
\end{figure}

\paragraph{Extension to 6-15 arithmetic steps}
We also extend the main GF(11) setting, where the RL dataset consists of problems with 6-9 arithmetic steps, to a more challenging setting where the RL data requires 6-15 steps of arithmetic reasoning (chosen uniformly). Here we only trained mixed-reasoning models. The results, displayed in~\cref{fig:ext_main,fig:ext_breakdown}, are similar to those in the main setting (\cref{fig:main,fig:passk_main}). We see that after RL training the model nearly saturates even this harder dataset and, notably, the performance on inversion problems reflects substantial strategy improvement. 

\section{Additional details}
\label{sec:additional-details}

All experiments in this paper were run on 1-4 A100-40GB GPUs. Each SFT run took 1-2 hours to complete. The RL experiments took 2-3 hours per run.

In \cref{fig:composition_example} we provide an example of a 2-part composition problem and its solution over $\text{GF}(11)$. The problem chains two inversion sub-problems, where the answer from Part 1 ($y_1$) is propagated to Part 2.

\begin{figure}[t]
\input{figs/composition_example}
\caption{Example of a 2-part composition problem over $\text{GF}(11)$ and its solution. The problem chains two inversion sub-problems, where the answer from Part 1 ($y_1$) is propagated to Part 2. The model must correctly solve both sub-problems to get the final answer.}
\label{fig:composition_example}
\end{figure}


\end{document}